\documentclass[conference]{IEEEtran}
\IEEEoverridecommandlockouts
\usepackage{graphicx}
\graphicspath{{figures/}}
\usepackage[caption=false,font=footnotesize]{subfig}
\usepackage{float}

\usepackage{cite}

\usepackage{amsmath,amssymb,amsfonts}
\usepackage{textcomp}
\usepackage{xcolor}
\usepackage{romannum}
\usepackage[linesnumbered,ruled,vlined]{algorithm2e}

\makeatletter
\newcommand{\removelatexerror}{\let\@latex@error\@gobble}
\makeatother
\usepackage{algpseudocode}

\SetKwInput{KwInput}{START}                
\SetKwInput{KwOutput}{Output}              

\def\BibTeX{{\rm B\kern-.05em{\sc i\kern-.025em b}\kern-.08em
    T\kern-.1667em\lower.7ex\hbox{E}\kern-.125emX}}

\begin{document}

\title{TextMage: The Automated Bangla Caption Generator Based On Deep Learning \\
}

\author{\IEEEauthorblockN{Abrar Hasin Kamal$^{1}$, Md. Asifuzzaman Jishan$^{2}$ and Nafees Mansoor$^{3}$}
\IEEEauthorblockA{Institute of Computer Science and Computational Science, Universität Potsdam, Germany$^{1}$}
\IEEEauthorblockA{Faculty of Statistics, Technische Universität Dortmund, Germany$^{2}$}
\IEEEauthorblockA{Department of Computer Science and Engineering, University of Liberal Arts Bangladesh$^{3}$}
Email:{ abrar.hasin.kamal@uni-potsdam.de$^{1}$, md-asifuzzaman.jishan@tu-dortmund.de$^{2}$, nafees@ieee.org$^{3}$}
}

\maketitle

\begin{abstract}
Neural Networks and Deep Learning have seen an upsurge of research in the past decade due to the improved results. Generates text from the given image is a crucial task that requires the combination of both sectors which are computer vision and natural language processing in order to understand an image and represent it using a natural language. However existing works have all been done on a particular lingual domain and on the same set of data. This leads to the systems being developed to perform poorly on images that belong to specific locales' geographical context. TextMage is a system that is capable of understanding visual scenes that belong to the Bangladeshi geographical context and use its knowledge to represent what it understands in Bengali. Hence, we have trained a model on our previously developed and published dataset named BanglaLekhaImageCaptions. This dataset contains 9,154 images along with two annotations for each image. In order to access performance, the proposed model has been implemented and evaluated. 
\end{abstract}


\section{Introduction}
Generating captions for images automatically through combination of computer vision and NLP is widely known as the process of Image Captioning in the field of intelligent systems. Image captioning tasks have been deemed much complex than object recognition tasks, that have been explored extensively in the past \cite{b1}. 

In recent years, automated image captioning using deep learning has received noticeable attention which resulted in the development of various models that are capable of generating captions in different languages for images \cite{b2}. We can understand that to produce captions in different languages each model has to be tailored differently\cite{b16}. Prior studies focused on methods that divided the task into two sub-tasks and solved the problem by stitching the results together at the end. More recently, a single joint model has shown to deliver better results. We identified that majority of the existing systems have a western bias, both contextual and lingual \cite{b5}.  

In this paper, we propose an automated image describing system that follows the conventional procedure of extracting features from an image using a CNN and feeding its preceding output layer to a language generating RNN consisting of LSTM cells to produce a caption in Bangla that describes the image. A dataset having a south Asian bias has been prepared for the task. A total of 9,154 images and two corresponding human annotations in the Bangla language for each image has been prepared for the problem. The dataset is available in the dataverse of Mendeley \cite{b7}. We first look at pre-existing work in the domain before defining the methodology of our paper. We then show the results of our model followed by the discussion and finally the conclusion. 

In the interim, the essential goals and commitments of this paper are, first, to build a dataset that eliminates the existing bias problem. Secondly, to construct an image captioning model which can generate Bangla inscription from any given image. And lastly, effectively testing the proposed model against predefined evaluation metrices for image captioning tasks. 

\section{Recent Work}
Image classification refers to the process of classifying objects from images of a dataset.Hence, to improve accuracy, the Inception-v3 implementation of the convolutional neural network is hailed as the best technique for image classification according prior literature. We use this as the philosophy for the classification of images in the image processing part of our work. Several papers reported the use of stochastic slope plunge streamlining alongside min-cut based calculation to position the boundaries over the images \cite{b19}. To rival this approach, the Hyper-spectral imaging (HSI) strategy has been proposed for delivering superior exactness and grouping capabilities while reducing computational time in another paper \cite{b23}.

Convolutional Neural Networks, also known as CNNs or Convnets, are the type of neural networks that are specialized in visual image analysis. CNNs are used in most cases for extracting the feature vector from an image in almost all of the past works that have been done till now \cite{b15}. Convolutional networks consist of a number of layers that process information and propagate them forward. At the top of the CNN, fully connected layers are responsible for flattening the output from the convolutional layers into a vector output. The data in each layer of the CNN is a three-dimensional array that is of size height x width x feature/color channel dimension \cite{b21}.

Furthermore, the Recurrent Neural Network are developed to deal with tasks that require the handling and processing of time-series data. RNNs were initially very resource heavy and problematic to train because of the back propagated gradients growing or shrinking. The developments in the architecture of RNNs now have made them an appropriate neural net for predicting what the next character or word may be in a sequence\cite{b24}.

In order to correct this problem of RNNs, the LSTM (Long Short-Term Memory) architecture was introduced in 1997 by Hochreiter and Schmidhuber. The primary objective of the model was to tackle the vanishing gradients issue. The architecture of LSTMs is very similar to that of a simple RNN. The only difference being the augmentation of an explicit memory cell in LSTMs \cite{b24}. The ordinary nodes in the hidden layer of RNNs is replaced by the memory cell in LSTM, which is consists of a node that is connected to itself via a recurrent edge. This recurrent edge has a constant weight of 1. The constant weight of this recurrent edge fulfills the lacking of a standard RNN which are the vanishing gradient problem \cite{b23}\cite{b24}.

Generating text from the image is the most pivotal task where the  thought process is to produce picture inscription using a complex neural network model. Moreover, utilizing Attention Generative Adversarial Networks proves to be the best method and accomplishes state of the art results \cite{b24}. Then again, one paper demonstrated that making and building up another model which combines CNN, RNN and LSTM model is working well and shows promising results \cite{b27}. FGGAN is another in demand exploration to illuminate picture subtitling strategy and it has a decent effectiveness which can create text from obscure or low-quality picture \cite{b27}. Moreover, they additionally demonstrated that text generation strategy relies upon the information effectiveness, information resize, and furthermore size of the dataset. Moreover, BLEU and METEOR assessment are significant subjects for the evaluation of models. They accomplished assessment scores of 63.5 and 30.6 for the human exhibition utilizing remarkable results for the three different dataset which is the flickr8K, flickr30k and the MS COCO.

\section{Dataset}
Dataset is a crucial part of the generating Bangla text from the given input image task and we have given considerable attention towards the tailored dataset which is named, BanglaLekhaImageCaptions. In like manner, the images amassed together has been built up as the most unique and precise in the innovative work of complex information-driven application frameworks as of late. The dataset contains  9,154 images and all images are kept in the same format. We included two annotations in Bangla for each image in this dataset.  
\begin{figure}[htbp]
\centerline{\includegraphics[scale=0.6]{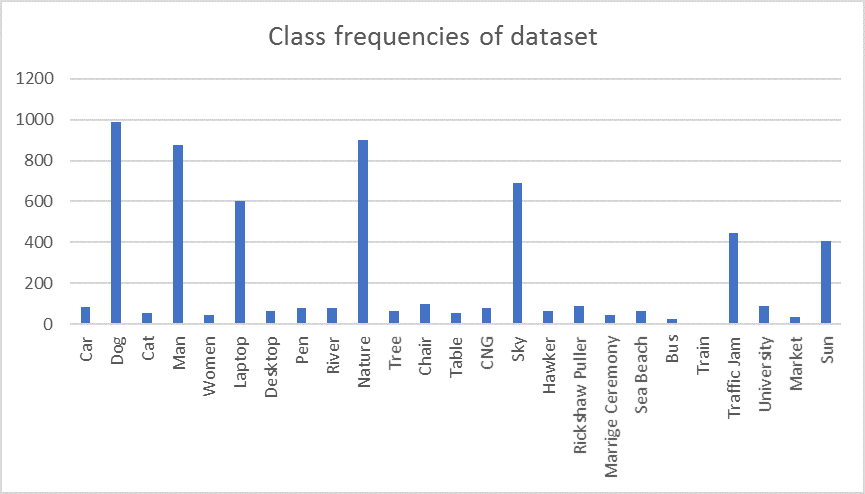}}
\caption{Class frequency.}
\label{fig1}
\end{figure}

\begin{table}[htbp]
\caption{Technical characteristics of BanglaLekhaImageCaptions dataset}
\begin{center}
\begin{tabular}{|c|c|c|c|}
\hline
\textbf{}&\multicolumn{3}{|c|}{\textbf{Resizing and Formation}} \\
\cline{2-4} 
\hline
No of images & File Format & Image dimension (H X W) & Bit depth  \\
\hline
9,154 & PNG & 224 × 224 & 24
\end{tabular}
\label{tab2}
\end{center}
\end{table}

While learning a machine, the bigger the dataset, the better the accuracy to produce results. To deal with the repository of 9,154 images, we mapped it into 25 classes. For the clarification of images, two annotations for each image have been included and images were resized similarly as they were not uniformly sized initially. Here, all images files are store in the PNG file format. Preceding the CNN usage, we resized the full dataset into 224 × 224 pixels and kept all 3 RGB channels for information retention. Figure 1 represents distribution of the classes for our dataset and Table 1 shows the technical description of images in the dataset. 

\section{Methodology}

In this section, we briefly look at the setup of the model and the optimization criterion that have been selected for the model.

\subsection{Simulation setup}
First, we give concern to the resizing portion of our dataset. We resize the whole dataset to 224 x 224 x 3 pixels. We utilized the VGG-16 features where the 1000 pre-trained class for the CNN pre-exist. In the CNN implementation, we found our dataset to comprise of 25 classes. The CNN architecture consisted of Conv2D, Maxpooling2D, ReLU activation, Flatten and Fully connected (FC) layers that gave us the feature vector for the next portion of the task. Maxpooling2d has been used for downsampling the images and reducing the dimensions that allows us the model to make assumptions about existing features of the image.  The flatten layer is responsible for flattening the input matrix into a vector while the FC layers are responsible for the final feature vector that we receive. 

Secondly, we were concerned about the RNN and LSTM, generally, used to create textual description from given data images. We included two Bangla annotations for every image. 256 channels in the LSTM were used and dropout esteem was set to 0.5. At long last, we led a completely associated layer with the Adam enhancement procedure, picked batch size 128, and estimated loss using cross-entropy. After training the CNN model we chose the best weight from the training period and generated an hdf5 weight model which has been used in the 25 epochs during the RNN training.

Finally, both the CNN and RNN models were stitched and trained on the dataset for 35 epochs again to generate a loss and accuracy graph. After completion of the training, all readied models were taken saved in our CV folder. Meanwhile, we adopted the preparations to evaluate our trained model for the mentioned dataset for indications of progress.

\subsection{Optimization}
Stochastic gradient descent (SGD) has been used for optimization in CNN with a batch size of 16 images. Besides, it can fix on a minima quicker than other optimization strategies and refresh information all the more regularly. We implemented the learning rate of 0.01, decay rate 1e-6, along with the rate of momentum=0.7, and nesterov=True since a dataset containing 9,154 pictures was used.

In the RNN part, we utilized the Adam optimization technique for the images to Bangla language automatic caption generator using a neural network. Adam optimizer mainly is a combination of RMSprop technique and stochastic gradient descent technique along with momentum. We have also concerned and maintained the accuracy and losses in the RNN period.

\section{Results and Discussion}
In this segment, we represent about the consequences of our model on the previously mentioned dataset. We focused mainly on training and validation of the model and used accuracy and loss as means of measure. Furthermore, BLEU and METEOR, which are widely known evaluation metrics for NLP have been used to evaluate our results.  
\subsection{Implementation of Convolutional Neural Network}
\begin{figure}[htbp]
\centerline{\includegraphics[scale=0.58]{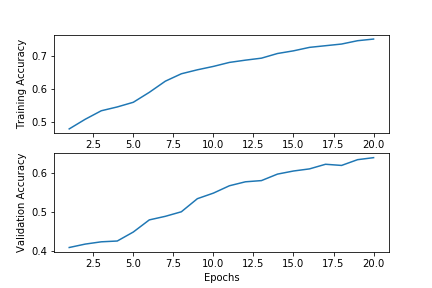}}
\caption{Image characterization result of CNN part.}
\label{fig2}
\end{figure}

Initially the research concerns towards CNN for image classification where, we classified the full dataset using 25 types of classes. We used the SGD optimization technique for the classification and ran 20 epochs for our self-made Bangla dataset. During the first epoch, we got better accuracy and after completing the training we achieved 0.758565 for the train set and 0.643476 for the validation purpose. We represented the training and validation time accuracy in \ref{fig2}. 
\subsection{Implementation of Recurrent Neural Network}
Next, we concerned regarding the RNN portion of our dataset. For the language part, we implemented 25 epochs during RNN training with selected batch size of 128. We represented the result of RNN which is 0.807854 in Figure 3 where we showed that the training time accuracy result and loss result of the RNN portion. Each image in the dataset has two corresponding human Bangla language annotations which resulted in 18,308 annotations for the full dataset. During training the loss function increased when the step size overstepped in the minima and therefore after that we can see from Figure 3 that it gradually decreased and reached a minimum after 25 epochs.

\begin{figure}[htbp]
\centerline{\includegraphics[scale=0.30]{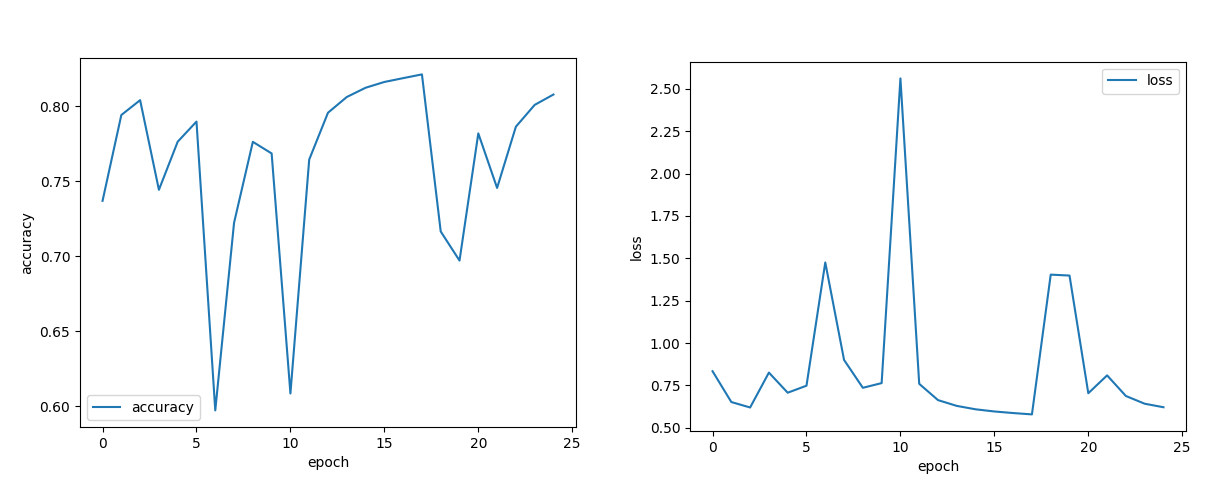}}
\caption{Training time accuracy and loss for RNN part.}
\label{fig3}
\end{figure}

\subsection{Image caption generating model implementation}
Combining the CNN and RNN that has been discussed, we implemented the complete image to Bangla caption generating model which is capable to generate captions from the given input image. For the final stage, we ran 35 epochs and each epoch took 2 hour 40 minutes to complete the training. The model showed approximately 0.916543 accuracy for training and 0.739776 for the validation period as seen in Figures 4 and 5.  
\begin{figure}[htbp]
\centerline{\includegraphics[scale=0.50]{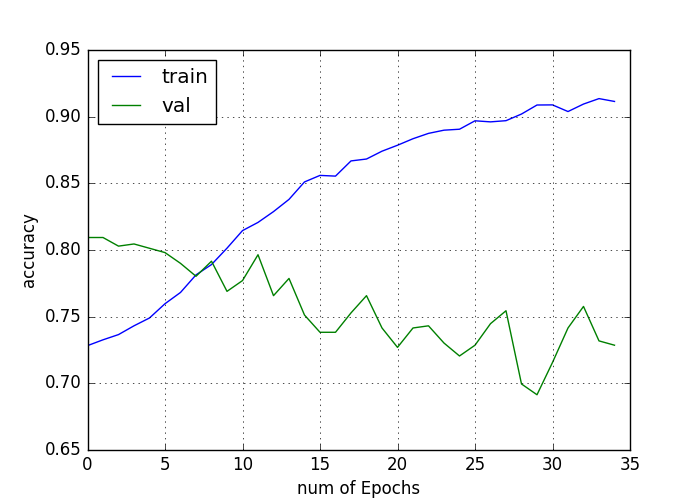}}
\caption{Training accuracy and validation accuracy for the final train up.}
\label{fig}
\end{figure}
\begin{figure}[htbp]
\centerline{\includegraphics[scale=0.50]{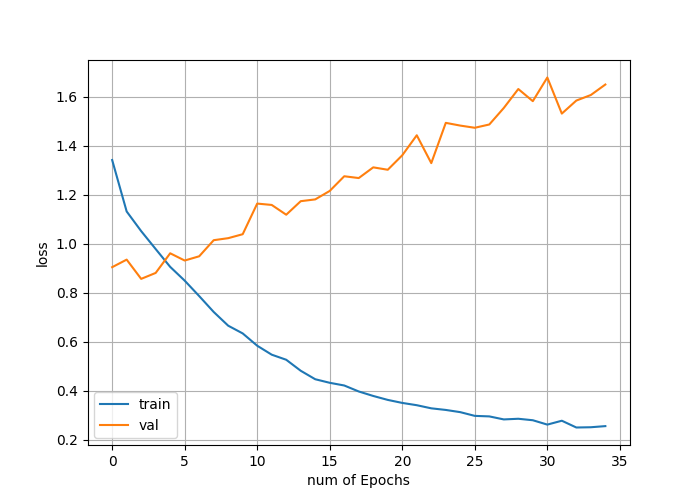}}
\caption{Training time loss and validation time loss for the final train up.}
\label{fig}
\end{figure}
\subsection{Evaluation Result}
After training the model, we evaluate our model using the BLEU and METEOR evaluation metric system. There are various sorts of assessments in the NLP and computer vision verse, however, among them, METEOR (Metric for Evaluation of Translation with Explicit Ordering) and BLEU (bilingual evaluation understudy) assessment are regarded as good measures for language related task evaluation. Moreover, we implemented BLEU-1, 2, 3, 4 along with the METEOR metric evaluation system. We used the different hidden layer sizes and they are the 32, 64, 128 respectively, and represented their evaluation scores in Table 2. 

We evaluated the BLEU-1, 2, 3, 4 scores and METEOR scores for the BanglaLekhaImageCaptions dataset and compared our results with the benchmark results of Multi-modal RNN \cite{b27}, Flickr8K \& Google NIC \cite{b2}, and LRCN \cite{b6} models which research result based on the Flickr8K, Flickr30K, and MS COCO dataset and showed the benchmark result for that dataset in Table 3. Moreover, we compared our benchmark result with that mentioned existing work because we implemented a new task. Our research-based on the image to Bangla language caption generation which is a unique work because we implemented our proposed model for our self-made image dataset. One restrictions of other model is that they can't produce distinctive example of sentence acknowledge as the datasets comprises of handcrafted comments, however our model can create dynamic yield as our model figures out how to regulate the extent of the area and word inserting.
\begin{table}[htbp]
\caption{BLEU and METEOR evaluation score for BanglaLekhaImageCaptions dataset}
\begin{center}
\begin{tabular}{|c|c|c|c|c|c|}
\hline
\textbf{}&\multicolumn{3}{|c|}{\textbf{Evaluation Score }} \\
\cline{2-4} 
\hline
Hidden layer & BLEU-1 & BLEU-2 & BLEU-3  & BLEU-4 &METEOR\\
\hline
32 & 66.7 & 43.6 & 31.5  & 23.8 & 18.227456\\
\hline
64 & 64.1 & 41.1 & 31.7  & 22.7 & 17.906543\\
\hline
128 & 65.8 & 43.5 & 30.9  & 23.1 & 18.458765\\
\end{tabular}
\label{tab1}
\end{center}
\end{table}

\subsection{Discussion}
We constructed an image captioning model capable of generating captions for images in Bangle. The proposed model is the mix of CNN, RNN, and LSTM model and is utilized on the independent BanglaLekhaImageCaptions dataset. We accomplished better accuracy, and it exhibited its capability to produce text from a given picture. Also, we accomplished remarkable results in the CNN part for our dataset (0.758565 for the training period and 0.643476 for the validation). A batch size of 16 was used and executed the SGD optimization method. We then dedicated our concerns towards the RNN part and achieved 0.807854 during training with the implemented Adam streamlining method. At last, we combined the two models and prepared the full framework for an execution of 35 epochs where our accuracy reached 0.916543 in training and 0.739776 for validation. In Table 2 and 3, we represented our BanglaLekhaImageCaptions dataset assessment test score on the BLEU-1, BLEU-2, BLEU-3, BLEU-4, and METEOR metrices along with the comparable benchmark result of Flickr8K, Flirckr30K, MS COCO in the field of image to text generation. 

\begin{table}[htbp]
\caption{BLEU and METEOR score for existing three benchmark dataset}
\centering
\begin{tabular}{c l c c c c c}
\hline 
Dataset  & Model            & B-1 & B-2 & B-3 & B-4 & METEOR    \\
\hline 
         & Multi-modal RNN \cite{b27}      & 58     & 28     & 23     & -      & -         \\
Flickr8K & Google NIC\cite{b2}      & 63     & 41     & 27     & -      & -         \\
    & LRCN \cite{b6}            & -      & -      & -      & -      & -         \\
\hline

\hline 
         & Multi-modal RNN \cite{b27}       & 55     & 24     & 20     & -      & -         \\
Flickr30K         & Google NIC \cite{b2}       & 66.3     & 42.3     & 27.7     & 18.3      & -         \\
 & LRCN \cite{b6}            & 58.8      & 39.1      & 25.1      & 16.5      & -         \\
\hline

\hline
         & Multi-modal RNN \cite{b27}      & -     & -     & -     & -      & -         \\
MSCOCO        & Google NIC \cite{b2}        & 66.6     & 46.1     & 32.9     & 24.6      & -         \\
 & LRCN \cite{b6}             & 62.8      & 44.2      & 30.4      & -      & -         \\
\hline 
\end{tabular}
\end{table}

\section{Conclusion}
In this paper we have presented an automated image captioning system, TextMage, that can perceive an image with a south Asian bias and describe it in Bangla. The model constructed for TextMage was heavily inspired from the first joint model "Show and Tell: A Neural Image Caption Generator" from an architecture perspective. Using our dataset of 9,154 images and 18,308 human annotations in Bangla, our system showed a validation accuracy of 73.98 and a training accuracy of 91.65. Using the dataset that has been used in this paper and published, future works can include more newer methods (e.g. Attention over LSTM) for benchmark results.

\end{document}